\documentclass[runningheads]{llncs}
\usepackage{graphicx}
\usepackage{bbm}
\usepackage{CJK}
\usepackage{hyperref}       % hyperlinks
\usepackage{algorithm}
\usepackage{algorithmicx}
\usepackage{amsmath}
\usepackage{algpseudocode}
\usepackage{threeparttable}
\usepackage{amsfonts}       % blackboard math symbols
\usepackage{subfigure}
\begin{document}
\title{Learning Optimal Treatment Strategies for Sepsis Using Offline Reinforcement Learning in Continuous Space}
\titlerunning{Sepsis AI}
\author{Zeyu Wang\inst{1} \and
Huiying	Zhao\inst{2}\and
Peng Ren\inst{3}\and
Yuxi Zhou\inst{3} \and
Ming Sheng\inst{3}} 
\authorrunning{Wang et al.}

\institute{Beijing Institute of Technology, Beijing, 100081, China\\ 
\email{wangzeyu@bit.edu.cn}\\ \and
Peking University People's Hosipital, Beijing, 100044, China\\
\email{zhaohuiying109@sina.com}\\ \and
BNRist, DCST, RIIT, Tsinghua University, Beijing, 100084, China\\
\email{\{renpeng, yuxi, shengming\}@tsinghua.edu.cn}
}
\maketitle              % typeset the header of the contribution
\begin{abstract}
Sepsis is a leading cause of death in the ICU. It is a disease requiring complex interventions in a short period of time, but its optimal treatment strategy remains uncertain. Evidence suggests that the practices of currently used treatment strategies are problematic and may cause harm to patients. To address this decision problem, we propose a new medical decision model based on historical data to help clinicians recommend the best reference option for real-time treatment. Our model combines offline reinforcement learning and deep reinforcement learning to solve the problem of traditional reinforcement learning in the medical field due to the inability to interact with the environment, while enabling our model to make decisions in a continuous state-action space. We demonstrate that, on average, the treatments recommended by the model are more valuable and reliable than those recommended by clinicians. In a large validation dataset, we find out that the patients whose actual doses from clinicians matched the decisions made by AI has the lowest mortality rates. Our model provides personalized and clinically interpretable treatment decisions for sepsis to improve patient care.

\end{abstract}

\section{Introduction}
Sepsis is a severe infection that can result in life-threatening acute organ dysfunction and is known as the leading cause of death in critically ill patients \cite{singer2016third}. It affects more than 49 million people around the world each year, killing between one in six and one in three of those affected \cite{rudd2020global}\cite{fleischmann2020incidence}\cite{rhee2017incidence}. Early effective resuscitation and haemodynamic management are crucial for the stabilisation of sepsis-induced tissue hypoperfusion in sepsis and septic shock, and they are more important for the prognosis \cite{rhodes2017surviving}\cite{lat2021surviving}. Although the Surviving Sepsis Campaign (SSC) guidelines 2021 recommend an initial target mean arterial pressure (MAP) of 65 mmHg \cite{rhodes2017surviving}, the following questions are not answered: 1) what is the optimal dose of fluid and how should it be titrated? 2) what is the optimal approach to selection and dose titration for vasopressor therapy? 3) which patients should glucocorticoid therapy be initiated for? To resolve these concerns, it is essential to carry out personalized therapies in real time based on the individual characteristics and status of patients.

In previous study, high-granularity dataset and reinforcement learning approach were adopted to explore the sequential role of the therapy strategy \cite{komorowski2018artificial}. However, its action and state are based on discrete space and there is a lack of more refined guidance for the treatment received by patients. Therefore, in our work, we propose a model to make medical decisions for sepsis patients based on historical data. We model in a continuous state-action space, representing the physiological state of a patient at a point in time as a continuous vector. LSTM mechanism is applied to capture the historical information of treatment received by the patient. In addition, offline deep reinforcement learning methods are used to determine the optimal treatment strategy. Finally, we conduct experiments to demonstrate that the strategy recommended by the model outperforms the clinician’s strategy in terms of survival rate and safety rate. Also, we find out that the mortality rate of patients is the lowest when the clinician’s treatment strategy is similar to the recommended strategy of the model.

Our contributions are as follows. We have introduced the offline reinforcement learning algorithms to better address the inability to interact with the environment in the medical field. The deep reinforcement learning models with continuous state-action spaces are implemented, and the optimal strategies are learned to improve patient outcomes and reduce patient mortality. We design experiments on the Medical Information Mart for Intensive Care version IV (MIMIC-IV) dataset to validate the model. The results show that the survival and safety rates of sepsis patients are significantly improved. At the same time, the analysis of the results reveals that the current method of drug use can be optimized, which is a guidance for the treatment of sepsis.

\section{Related Work}
Reinforcement learning approaches have been extensively explored in the treatment of patients with severe sepsis.

In discrete space, the Fitted-Q Iteration algorithm\cite{r2} was applied to learn treatment strategies for mechanical ventilation weaning from historical data\cite{r1}; Komorowski et al.\cite{komorowski2018artificial} discretized the state and action space through k-means clustering, and then performed Q-learning \cite{sutton2018reinforcement} to generate the optimal strategy of managing intravenous fluids and vasopressors.

In continuous space, Raghu et al.\cite{raghu2017deep} used Dueling Double-Deep Q Network(\cite{wang2016dueling}, \cite{van2016deep}) to learn medical treatment policies for sepsis. This approach used a vector representation of continuous states to extend the treatment of sepsis to a continuous space. Sun et al.\cite{sun2021personalized} combined reinforcement learning and supervised learning, with the DDPG method adopted to develop strategies in a continuous value space.

In this work, we also focus on the treatment of sepsis, but aim to develop a model that does not interact with the environment in a continuous state-action space. In turn, it solves the performance problem of reinforcement learning in the medical field that it cannot do the exploration, while optimizing the treatment process. Additionally, more refined medical actions are taken.

\section{Preliminaries}
\subsection{Reinforcement Learning}
In reinforcement learning, time series data are often modeled with Markov Decision
Processes (MDP)$(S,A,p_{M},r,\gamma)$, with state space $S$, action space $A$, and transition dynamics $p_{M}(s_{0} |s, a)$. At each discrete time step, the agent performs action $a$ in the state $s$ and arrives at the state $s'$, while the agent receives a reward $r \in R$. The agent selects the action to maximize the expected discounted future reward, known as the return defined as $R_{t} = \sum_{t'=t}^{T}\gamma^{t'-t}r_{t'}$, where $\gamma \in (0,1)$, represents the discount factor, capturing the tradeoff between immediate and future rewards and $T$ refers to the terminal timestep. The agent selects action according to a policy $\pi : S \rightarrow A$. And each policy $\pi$ has a $Q$ function $Q^{\pi}(s,a) = \mathbb{E}_{\pi}[R_{t}|s,a]$. For a given policy, the Q function can be computed using the Bellman equation:
\begin{equation}
Q^{\pi}(s_{t},a_{t}) = \mathbb{E}_{r_{t},s_{t + 1} \sim E}[r(s_{t},a_{t}) + \gamma \mathbb{E}_{a_{t + 1} \sim \pi}[Q^{\pi}(s_{t + 1},a_{t + 1})]]
\end{equation}
And if the target policy is deterministic, we use the policy directly:
\begin{equation}
Q^{\pi}(s_{t},a_{t}) = \mathbb{E}_{r_{t},s_{t + 1} \sim E}[r(s_{t},a_{t}) + \gamma Q^{\pi}(s_{t + 1},\pi(s_{t + 1}))]
\end{equation}
We consider continuous state-action space model-free RL and use historical data to find a good-quality policy $\pi$.
\subsection{Extrapolation Error}
As for reinforcement learning tasks in the medical field, it has to learn from historical data because of the high cost incurred by the interaction between agent and environment. This may lead to extrapolation errors. We define $\epsilon_{MDP}$ as the extrapolation error. This accounts for the difference between the value function $Q^{\pi}_{\mathcal{B}}$ computed with the history data and the value function $Q^{\pi}$ computed with the environment:
\begin{equation}
\epsilon_{MDP}(s,a) = Q^{\pi}(s,a) - Q^{\pi}_{\mathcal{B}}(s,a)
\end{equation}
Such errors will cause an even greater problem in continuous state space and multidimensional action space. Avoiding extrapolation errors plays a critical role in ensuring safe and effective patient care. Fujimoto et al.\cite{fujimoto2019off} relied on batch-constrained reinforcement learning to solve this problem well. Additionally, Fujimoto et al.\cite{fujimoto2019off} demonstrated that the extrapolation error can be eliminated and that BCQL can converge to the optimal policy on this MDP corresponding to dataset $\mathcal{B}$.

\section{Datasets}
Our experimental data are obtained from the Multiparametric Intelligent Monitoring in Intensive Care (MIMIC-IV) database. We focus on those patients who met sepsis-3 criteria \cite{singer2016third} (6660 in total) within the first 24 hours of admission to the hospital. Sepsis is defined as a suspected infection (prescription of antibiotics and sampling of bodily fluids for microbiological culture) combined with the evidence of organ dysfunction, as defined by a SOFA score $\geq$ 2 within 24 hours of admission. In line with previous research, we assume a baseline SOFA of zero for all patients \cite{lat2021surviving}\cite{seymour2016assessment}\cite{raith2017prognostic}\cite{seymour2019derivation}. For each patient, we have the relevant physiological parameters, including demographics, comorbidities, vital signs, laboratory values, treatment interventions, intake/output events and 90 day mortality.

Since the first 24 hours are extremely critical for the treatment of sepsis, we extract data within 24 hours of patient onset. The data are aggregated into 2-hour windows. Besides, when there are several data points in a window, the average or sum (as appropriate) is recorded. This produces a $41\times1$ feature vector for each patient at each time period, which is the state $s_{t}$ in the base MDP. \\
\\
The physiological features used in our model are as follows:\\
\textbf{Demographics:} gender, age, ethnicity; \\
\textbf{Comorbidities:} elixhauser premorbid status;\\
\textbf{Vital signs:} heart rate, mean arterial pressure (MAP), temperature, respiratory rate, peripheral capillary oxygen saturation (SpO2), glasgow coma scale (GCS); \\
\textbf{Lab values:} white blood cell count (WBC), neutrophils, lymphocytes, platelets, hemoglobin, alanine aminotransferase (ALT), aspartate aminotransferase (AST), total bilirubin, blood urea nitrogen (BUN), creatinine, albumin, glucose, potassium, sodium, calcium, chloride, potential of hydrogen (PH), partial pressure of oxygen (PaO2), partial pressure of carbon dioxide (PaCO2), bicarbonate, PaO2/FiO2 ratio, lactate, prothrombin time (PT), activated partial thromboplastin time (APTT); \\
\textbf{Organ function score:} sequential organ failure assessment (SOFA) score; \\
\textbf{Output events:} urine volume; \\
\textbf{Treatment interventions:} 1) intravenous fluids volume; 2) the maximum dose of vasopressors: norepinephrine, phenylephrine, vasopressin, angiotensinii, epine-phrine, dopamine, dobutamine; 3) whether hydrocortisone was used; \\

\section{Model Architecture}
Our model architecture consists of four main components: History capture model, Generative model, Perturbation model and Q-networks. By using this model, the offline reinforcement problem of optimal decision-making in continuous stateaction space is effectively solved.
\subsection{History Capture Model}
The goal of history capture model is to capture the change of states while incorporating the influence of the performed action over time. In the history capture model, the observation-action history is explicitly processed by an LSTM network and fed as input into other networks. For each moment of the patient’s state, we use the historical treatment process $(\{o_{1},a_{1}\},...,\{o_{t},a_{t-1}\})$ as the input of the LSTM for calculation. Also, we will get an embedding representation $s_t$ of the patient’s current status by combining historical status and treatment information.
\subsection{Generative Model}
To avoid extrapolation error, a policy is supposed to induce a similar state-action visitation to the batch. The purpose of generative models as a model of imitative learning is to simulate the treatment strategies of clinicians by observing the state of the patient. By using this method, the model’s strategies are distributed over the range of the dataset.

For the generative model, we use a conditional variational auto-encoder (VAE)\cite{kingma2013auto}. The VAE $G_{\omega}$ is defined by two networks, an encoder $E_{\omega1}(s, a)$ and decoder $D_{\omega2}(s, z)$, where $\omega = \{\omega1, \omega2\}$. The encoder takes a state-action pair and outputs the mean $\mu$ and standard deviation $\sigma$ of a Gaussian distribution $N(\mu, \sigma)$. The state $s$, along with a latent vector $z$ as sampled from the Gaussian, is passed to the decoder $D_{\omega2}(s, z)$ which outputs an action. The VAE is trained with respect to the mean squared error of the reconstruction along with a KL regularization term:
\begin{equation}
\mathcal{L}_{VAE} = \sum_{(s,a) \in \mathcal{B}}(D_{\omega2}(s, z)-a)^{2}+D_{KL}(\mathcal N(\mu, \sigma)||\mathcal N(0, 1))
\end{equation}

\subsection{Perturbation Model}
To enhance the diversity of actions, we introduce a perturbation model $\xi_{\phi}(s,a,\varphi)$. The  perturbation model makes an adjustment based on action $a$ which is generated from the generative model in the range $[-\varphi,\varphi]$. In this way, the output of the model is restricted to the scope of the dataset. This results in the policy $\pi$:
\begin{equation}
\pi(s) = \mathop{argmax}\limits_{a_{i}+\xi_{\phi}(s,a,\varphi)}Q_{\theta}(s,a_{i}+\xi_{\phi}(s,a,\varphi)) , {a_{i}\sim G_{\omega}(s)}_{i=1}^{n}
\end{equation}
The perturbation model $\xi_{\phi}$ can be trained to maximize the $Q_{\theta}(s,a)$ through the deterministic policy gradient algorithm by sampling $a \sim G_{\omega}(s)$:
\begin{equation}
\phi \leftarrow \mathop{argmax}\limits_{\phi} \sum\limits_{(s,a)\in B} Q_{\theta}(s,a+\xi_{\phi}(s,a,\varphi))
\end{equation}
The choice of $n$ and $\varphi$ creates a trade-off between an imitation learning and reinforcement learning algorithm. If $\varphi = 0$ and $n = 1$, the model exhibits the characteristics of imitation learning, which imitates the clinician's strategy. And if $\varphi$ is unconstrained and $n \rightarrow \infty$, the model approaches DDPG(Deep Deterministic Policy Gradient), an algorithm which searches the policy to greedily maximize the value function over the entire action space.

\subsection{Q-networks}
Q-network is a method used to evaluate the value of a strategy with a neural network to approximate the value function. Deep Q-Network is an off-policy approach. Instead of using the real action of the next interaction for each learning, the target value function is updated by using the action currently considered to have the highest value. In this way, an overestimation of the Q value can occur. Clipped Double Q-learning estimates the value by taking the minimum between two Q-networks: $Q_{\theta1}$ and $Q_{\theta2}$. Also, taking the least operator also penalizes the high variance estimates in the uncertainty region and facilitates the action of strategy selection for the states contained in the dataset. In particular, we take a convex combination of the two values, with a higher weight on the minimum, to form a learning target which is used by both Q-networks:
\begin{equation}
r + \gamma\mathop{max}\limits_{a_{i}}[\lambda\mathop{min}\limits_{j=1,2}Q_{\theta_{j}'}(s',a) + (1 - \lambda)\mathop{max}\limits_{j=1,2}Q_{\theta_{j}'}(s',a)]
\end{equation}

Here is a summary of the model framework, which maintains four parametrized networks: a generative model $G_{\omega}(S)$, a perturbation model $\xi_{\phi}(s,a)$, and two Q-networks $Q_{\theta_{1}},Q_{\theta_{2}}$. In the meantime, each of the perturbation and Q-networks has 1 target network. Similar to the DQN method, the parameters of the target network are updated after a certain period of time.\\

\section{Experiment}
This section describes the training details for our models.
\subsection{Medical Action Selection}
An immediate action for resolving hypotension should be taken as quickly as possible for those sepsis patients with hypoperfusion. Fluid resuscitation and vasopressor management are essential for the treatment of hypotension and hypoperfusion. Norepinephrine and vasopressin are the first-line and second-line vasopressor, respectively. Inotropes such as dobutamine and norepinephrine are recommended to the patients with septic shock and cardiac dysfunction with persistent hypoperfusion. Glucocorticoids (first choice is hydrocortisone) are also recommended for refractory septic shock. Therefore, in the experiment, for the choice of medical behaviors, we divide them into three parts of refinement. The first part is the fluid input for patients every two hours. The second part is the use of antihypertensive drugs, in which we classify norepinephrine and phenylephrine as the first type of vasopressors, vasopressin and angiotensin II as the second type of vasopressors, and epinephrine, dopamine and dobutamine as the third type of vasopressors, according to pharmacological characteristics. In turn, we optimize the three classes of antihypertensive drugs. The third part is the use of hydrocortisone, which is a discrete type of decision-making behavior. The above three parts are most critical to the treatment of sepsis and are of great importance to clinical application.

\begin{table}[H]
\centering
\caption{The selection of medical actions}
\begin{threeparttable}
\begin{tabular}{cccc}
\hline
\textbf{Action} & \textbf{Content}     & \textbf{Unit}           & \textbf{Type}$^{\rm a}$ \\ \hline
liquid          & intravenous fluids              & milliliter/2h & continuous    \\
vasopressor\_1  & norepinephrine,phenylephrine    & microgramme/kg.min & continuous    \\
vasopressor\_2  & vasopressin,angiotensinii       & U/min & continuous    \\
vasopressor\_3  & epinephrine,dopamine,dobutamine  & microgramme/kg.min & continuous    \\
hydrocortisone  & hydrocortisone                  & - & discrete      \\\hline
\end{tabular}
\begin{tablenotes}
		\footnotesize
		\item a: Continuous type of action implies that we decide the specific value of the action. Discrete type of action implies that we decide whether to adopt the action or not.
\end{tablenotes}
\end{threeparttable}
\end{table}

\subsection{Reward Function}
For the design of the patient reward function, we integrate the intermediate treatment process of the patient with the final outcome. Since our goal is to provide guidance for patients within 24 hours after onset, we prefer to improve the change of patients’ status within 24 hours after onset. Therefore, for the change of patients in status, we consider a combination of two indicators, including the SOFA score and lactate level of patients. For the final outcome of the patient, we use the fact of whether the patient died while in the ICU as the final outcome.

Our reward function for intermediate timesteps is designed as follows:
\begin{equation}
r = C_{0}\,s_{t}^{SOFA} + C_{1}\,(s_{t+1}^{SOFA} - s_{t}^{SOFA}) + C_{2}\,tanh(s_{t+1}^{Lactate} - s_{t+1}^{Lactate})
\end{equation}
We conduct experiment with multiple parameters and opt to use $C_{0} = -0.1$, $C_{1} = -1$, $C_{2} = -2$

At terminal timesteps, we set a reward of +25 if a patient survived their ICU stay, and a negative reward of -25 otherwise.
\subsection{Training Process}
For our training process, our pseudocode is shown below. The details about our specific implementation can be found in our project code \url{https://github.com/taihandong-330/BCADRQN}.
 \begin{algorithm}[H]
    \caption{Batch-Constrained Action-specific Deep Recurrent Q-Network}
        \begin{algorithmic}[1] %每行显示行号
            \Require Records buffer $\mathcal{B}$ – observations $O$, actions $A$, reward function $R$;
        
            Parameters –  target network update rate $\tau$, mini-batch size $N$, max perturbation $\varphi$, number of sampled actions $n$, minimum weighting $\lambda$, number of epochs $M$;
            \State Randomly initialize LSTM data processing net $L$, with parameter $\psi$
            \State Randomly initialize VAE $G_{\omega} = \{E_{\omega1}, D_{\omega2}\}$, with parameter $\omega$
            \State Randomly initialize main perturbation net $\xi_{\phi}$, with parameter $\phi$
            \State Randomly initialize main Crisis net $Q_{\theta_{1}}, Q_{\theta_{2}}$, with parameter $\theta_{1}, \theta_{2}$
            \State Target perturbation net $\xi_{\phi}$: $\phi' \leftarrow  \phi$
            \State Target critic net $Q_{\theta_{1}'},Q_{\theta_{2}'}$: $\theta_{1}' \leftarrow  \theta_{1}, \theta_{2}' \leftarrow  \theta_{2}$
            \For{$m = 1 \to M$}:
                \State Initialize the batch buffer $\mathcal{D}$
                \For{$i = 1 \to N$}:
                    \State Initialize the first action $a_{0} = no \ operation$
                    \State Randomly select a patient at a time point $t$, sample a historical treatment episode $\langle (\{o_{1},a_{0}\},... \{o_{t},a_{t-1}\}, \{o_{t+1},a_{t}\}), r_{t}\rangle$ from $\mathcal{B}$
                    \State Store the historical treatment episode into $\mathcal{D}$
                \EndFor
                \State $s = L(\{o_{1},a_{0}\},... \{o_{t},a_{t-1}\}), s' = L(\{o_{1},a_{0}\},... \{o_{t+1},a_{t}\}), a = a_{t}, r = t_{t}$
                \State $\mu, \sigma = E_{\omega_{1}}(s,a)$,  $\tilde{a} = D_{\omega_{2}}(s,z)$, $z \sim \mathcal N(\mu, \sigma)$ 
                \State $\omega, \psi \leftarrow argmin_{\omega,\psi }\sum(a-\tilde{a})^{2}+D_{KL}(\mathcal N(\mu, \sigma)||\mathcal N(0, 1))$
                \State Sample $n$ actions:$\{a_{i} \sim G_{\omega}(s')\}$
                \State Perturb each action:$\{a_{i} = a_{i} + \xi_{\phi}(s',a_{i},\varphi)\}_{i=1}^{n}$
                \State Set value target $y = r + \gamma\mathop{max}\limits_{a_{i}}[\lambda\mathop{min}\limits_{j=1,2}Q_{\theta_{j}'}(s',a) + (1 - \lambda)\mathop{max}\limits_{j=1,2}Q_{\theta_{j}'}(s',a)]$
                \State $\theta \leftarrow argmin_{\theta} \sum (y - Q_{\theta}(s,a))^{2}$
                \State $\phi \leftarrow argmax_{\phi} \sum Q_{\theta_{1}}(s,a+\xi_{\phi}(s,a,\varphi))$, $a \sim G_{\omega}(s)$
                \State Soft update target networks: $\theta_{i}' \leftarrow \tau \theta + (1 - \tau)\theta_{i}'; \phi' \leftarrow \tau \phi + (1 - \tau)\phi'$
            \EndFor
    \end{algorithmic}
\end{algorithm}

\section{Results}
\subsection{Result Analysis}
For the results of the model training, we show the distribution of the model's output strategies relative to the clinician's original strategy. Figure 1 shows the difference between the model and the clinician's fluid input and the three classes of vasopressor within 24 hours of patient onset.

After analysis we find out that for intravenous fluids, the model’s strategy is approximately the same as that of the clinicians. However, the proportion of patients receiving vasopressors is only 10.7\% and 11.8\% for the first and second two hours after the onset of sepsis, but these would have been 14.1\% and 13.7\% if the recommendation made by AI Clinician was followed. There are also significant differences in the doses of the three classes of vasopressors. We find out that for the first and second classes of vasopressors, the model tends to select larger dosages. While for the third class of vasopressors, the model tends to select smaller dosages compared to the clinicians. In addition, we analyze the proportion of hydrocortisone use on the test set, discovering that the model use is essentially the same as the use by the clinician.
\begin{figure}[H]
\centering
\subfigure[Intravenous fluids]{
\includegraphics[width=5cm]{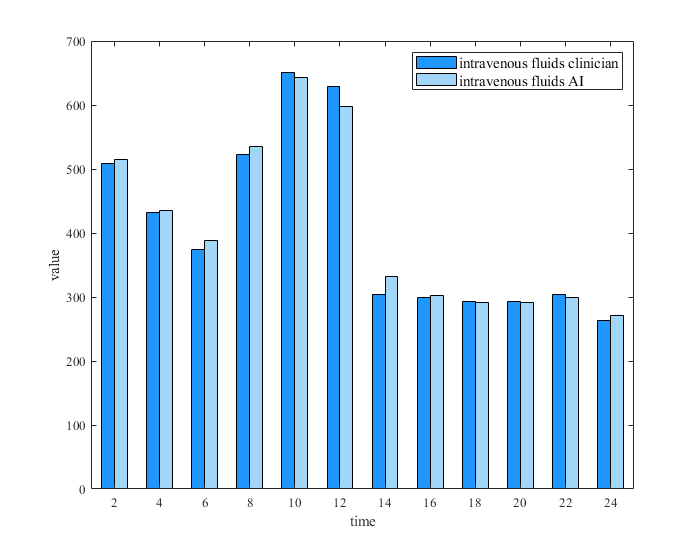}
%\caption{fig1}
}
\subfigure[Vasopressor]{
\includegraphics[width=5cm]{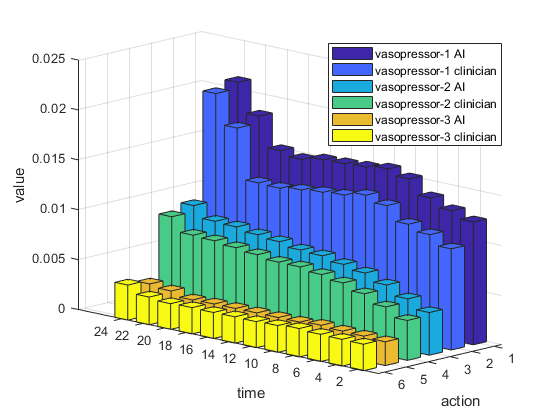}
}
\caption{The distribution of clinicians and AI strategies is shown for every two hours. The value of the strategy represents the average measure of all patients at the corresponding moment in time.}
\end{figure}
We further analyze the change in patient mortality when there is a difference between the clinician’s decisions and those of the model. We find out that, for the most part, patient mortality is lower when the clinician’s strategy differs from the model’s strategy insignificantly. Also, when the difference between the two is too large, the mortality rate of patients tends to increase substantially. This also demonstrates the validity of our model.
\begin{figure}[H]
\centering
\subfigure[Liquid]{
\includegraphics[width=2.6cm]{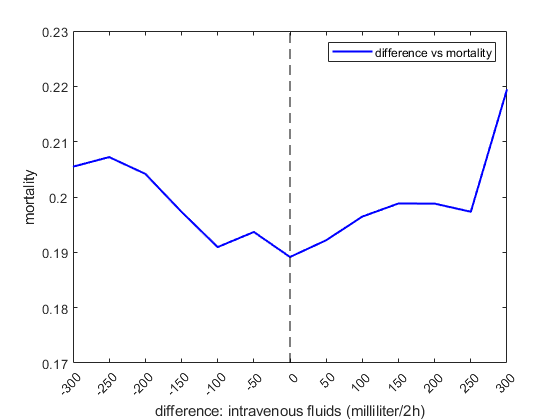}
%\caption{fig1}
}
\subfigure[Vasopressor-1]{
\includegraphics[width=2.6cm]{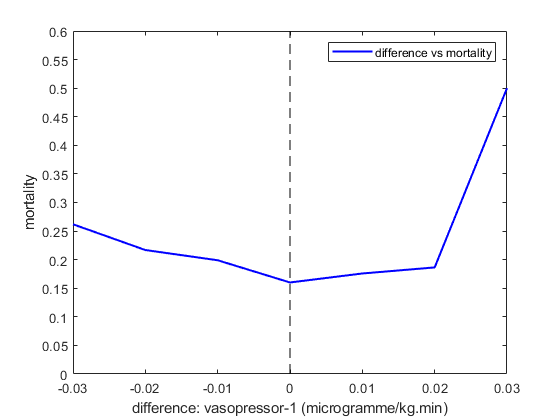}
}
\subfigure[Vasopressor-2]{
\includegraphics[width=2.6cm]{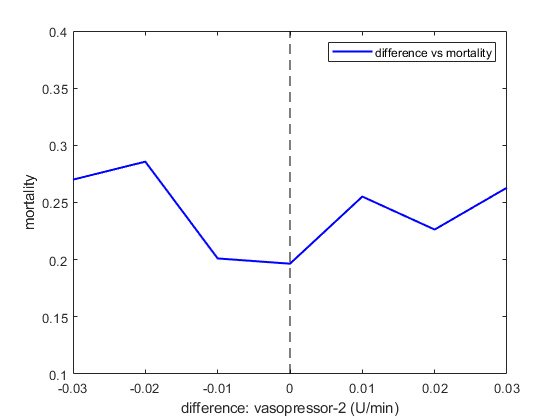}
}
\subfigure[Vasopressor-3]{
\includegraphics[width=2.6cm]{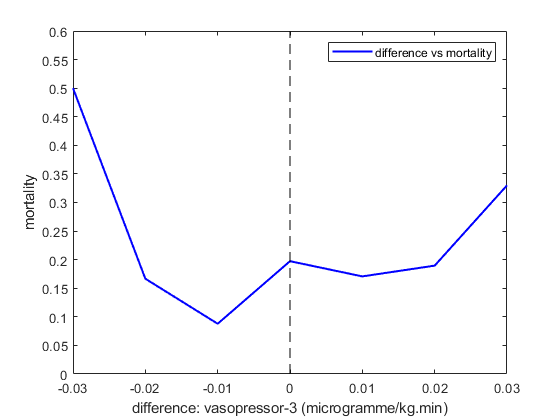}
}
\caption{Compare how mortality varies with the difference between the dose recommended by the optimal policy and the dose used by the clinicians. When the difference is smaller, we see lower observed mortality rates, suggesting that patient survival can be improved when clinicians act on the learned policy in AI.}
\end{figure}

\subsection{Evaluation Metric}
Since offline reinforcement learning is more difficult to measure in continuous space, this experiment focuses on two metrics for evaluation and the result is analyzed on the test set.
\subsubsection{Survival rate}
Improving patient mortality is particularly important in the healthcare process. Survival rate is an important metric for evaluating system performance. However, the offline reinforcement learning tasks in continuous space cannot interact with the environment to obtain rewards. Therefore, we use the Q function to evaluate the survival rate. 

\begin{figure}[H]
    \centering
    \includegraphics[width=6cm]{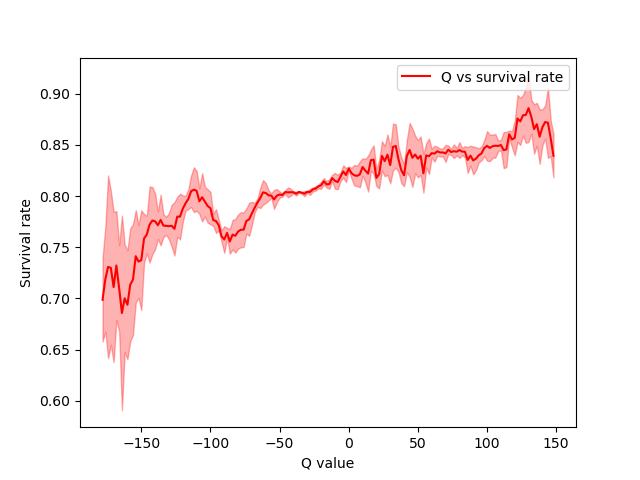}
    \caption{The relationship between Q and survival rate. The shadows are the result range of 5-fold cross-validation and there is a positive correlation between Q and the survival rate. Thus, a reasonable Q evaluation mechanism can be used to measure the result of the strategy in the offline case.}
    \label{Q}
\end{figure}

Q vs. survival rate links expected returns Q to survival rates. The survival rate of a Q value is:
\begin{equation}
survival\_rate\,(Q) = \dfrac{\#ofsurvival(s,a)_{Q_{i}}}{\#of(s,a)_{Q_{i}}}
\end{equation}

where the $(s,a)_{Q_{i}}$ means a state-action pair with $Q(s,a)\in {Q_{i}}$. ${Q_{i}}$ is an range of Q. 

In our experiments, we take a perturbation parameter $\varphi = 0.05$, which corresponds to a modified clinician-based strategy. Our experiments result in our Q value of 52.47, corresponding to a survival rate of 0.844, while the evaluated Q value of the clinicians' strategy is 13.19, corresponding to a survival rate of 0.813. This indicates that our model is optimized based on the clinicians' strategy.

\subsubsection{Safe rate}
Another evaluation metric under our consideration is the safety rate of the strategy. As for safety measures, we consider the AI-recommended drug doses in the range of 70\%-130\% of the clinician’s strategy to be safe.
\begin{equation}
safe\_rate = \dfrac{1}{N} \sum_{i=0}^N \dfrac{1}{T} \sum_{t=0}^T \bigcap^{a_{j}} \mathbbm{1}(0.7 < (\dfrac{V^{AI}_{a_{j}}}{V^{real}_{a_{j}}}) < 1.3)
\end{equation}

The final result of the safety rate for our experiments is 0.902, which means the safety of the model results is guaranteed to a large extent. In addition, the data quality issue affects our safety rate calculation to some extent.

\section{Conclusions}
In this paper, we implement an effective decision optimization system for sepsis treatment in a continuous decision space. The experimental results show that the optimized medical decisions can effectively improve the survival and prognosis of patients. This work makes several key contributions.

At the algorithm level, on the one hand, our algorithm introduces an offline reinforcement learning method, which is an effective solution to the extrapolation error in the offline environment. On the other hand, we capture the patient’s historical state, while extending the decision space to a continuous space, which is very important in reality.

At the medical level, our approach can well address the treatment of sepsis patients within 24 hours, improving their prognosis. We also refines the action of three kinds of vasopressor, fluid input, and hydrocortisone, which has more practical implications for optimizing clinicians’ decision.

Our analysis identifies that for intravenous fluids, the AI strategy is approximately the same as that of the clinician as well. Additionally, more fluid is required in the first 12 hours after the onset of sepsis. We also find out that such vasopressors as norepinephrine and vasopressin need to be early initiated and administered in larger doses. However, such inotropes as dobutamine and norepinephrine may require lower doses in sepsis treatment because of increased sympathetic stress and oxygen consumption. Finally, we also discover that compared to the real strategy of clinicians, no more patients are needed to receive glucocorticoid therapy.

In our future work, we will focus on improving more robust clinical reward mechanisms and constructing interpretable deep learning models. At the same time, we will continue generalizing them to a wider range of medical scenarios.

\bibliographystyle{splncs04}
\bibliography{reference}

\end{document}